\documentclass[11pt,a4paper]{article}
\usepackage{authblk}
\usepackage{hyperref}
\usepackage{changepage}
\usepackage{graphicx}

\begin{document}

\title{Do Chains-of-Thoughts of Large Language Models Suffer from Hallucinations, Cognitive Biases, or Phobias in Bayesian Reasoning?}
\author{Roberto Araya$^1$\thanks{roberto.araya.schulz@gmail.com}}
\date{%
    $^1$Institute of Education, CIAE, University of Chile\\
    \today
}
\maketitle

\begin{abstract}
Learning to reason and carefully explain arguments is central to students' cognitive, mathematical, and computational thinking development. This is particularly challenging in problems under uncertainty and in Bayesian reasoning. With the new generation of large language models (LLMs) capable of reasoning using Chain-of-Thought (CoT), there is an excellent opportunity to learn with them as they explain their reasoning through a dialogue with their artificial internal voice. It is an engaging and excellent opportunity to learn Bayesian reasoning. Furthermore, given that different LLMs sometimes arrive at opposite solutions, CoT generates opportunities for deep learning by detailed comparisons of reasonings. However, unlike humans, we found that they do not autonomously explain using ecologically valid strategies like natural frequencies, whole objects, and embodied heuristics. This is unfortunate, as these strategies help humans avoid critical mistakes and have proven pedagogical value in Bayesian reasoning. In order to overcome these biases and aid understanding and learning, we included prompts that induce LLMs to use these strategies. We found that LLMs with CoT incorporate them but not consistently. They show persistent biases towards symbolic reasoning and avoidance or phobia of ecologically valid strategies.

\end{abstract}



\section{\textbf{Introduction}} 

\noindent Given the sudden irruption of AI, developing the skills to communicate and collaborate with Artificial Agents (AA) is now more critical than ever. This means interacting with increasingly intelligent agents that will begin to surpass us in their ability to reason. In several problems, like in board games and biochemistry, they are already foreseeing several more steps than we are ~\cite{ananthaswamy2024close}, and simultaneously, they are improving their mind-reading skills ~\cite{strachan2024gpt}. Consequently, in several areas, they are starting to be able to make better decisions than us. These advances are great opportunities for improvements in productivity and well-being, but they also entail several risks. We can easily be manipulated, making us believe that we understand and support their decisions, which sometimes are controversial. This difference in intelligence is a tremendous existential risk. LLMs can make us profoundly convinced ~\cite{dennett2023problem} and induce a “deepity” effect. Coined by Daniel Dennett ~\cite{dennett2013intuition}, this effect refers to a feeling of understanding provoked by ambiguous ideas that seem important and trustworthy. Transcendence and triviality can quickly generate intrigue and pleasure cycles that attract and trap us ~\cite{Pinsof}. Subjective probability of events determined by specific characteristics of the population or by salient features is another well-known bias ~\cite{kahneman1972subjective}. We have many other cognitive biases ~\cite{van2022cognitive} that AA can use to hack us. One is our preference for arguments based on anecdotes and gossip rather than data and statistics ~\cite{Pinsof}. AAs can easily exploit these biases to achieve their goals.

One of the leading educational challenges is learning to reason critically. Students must learn to assess alternatives, balance with previous evidence, and carefully explain arguments. This is central to students' cognitive, mathematical, and computational thinking development. This challenge is particularly critical in problems under uncertainty. In these critical cases, students need to learn to take account of historical evidence with relevant priors, together with some indirect new evidence. This is Bayesian reasoning~\cite{pinker2022rationality}.

These types of problems are widespread, particularly in everyone's social life. Trust constantly arises when determining whether or not to trust others and AAs. Trust is a central element in social life, not only in interaction with others but in institutions and democracy. An illustrative example is the lyrics of the popular song Mother from Pink Floyd's The Wall:  "Mother, should I trust the government?"

Moreover, trust in non-kin and unknown people directly relates to commerce and economic and social development. Increased trust in strangers gave Europe an incredible advantage after the Middle Ages ~\cite{henrich2020weirdest}. The rise of trust in strangers decreases corruption and surges the exchange of ideas and innovation ~\cite{henrich2020weirdest}. When faced with a new interaction, there is, on the one hand, historical background, such as reputation or priors, but there is permanent doubt in the new situation. We do not know if we are now being given false information, lied to, or manipulated. Given the increasing intelligence of AAs that allows them to take better advantage of our limited rationality and cognitive biases, the trust conundrum is much more compelling and complex.

A significant additional challenge in developing Bayesian thinking in students is that in problems that touch on people's important beliefs, such as in politics, people with higher numerical ability use their quantitative reasoning capacity selectively to adjust their interpretation of the data to the result most consistent with their perspectives and beliefs ~\cite{kahan2017motivated}. Not always people are aware that they are doing this, leading them to make erroneous estimations, misjudgments, and poor decisions.   

Therefore, we need to develop a sharper rationality in our students than has been necessary until now. Our students must develop mathematical reasoning skills ~\cite{araya2022feasible} ~\cite{araya2023and} ~\cite{araya2021enriching}, statistical thinking  ~\cite{estrella2022early},  and computational reasoning skills ~\cite{araya2023unplugged}, capable of detecting inconsistencies in more complex arguments and longer chains of reasoning. This is\ particularly needed in situations where uncertainty is present. This should be part of AI Literacy initiatives, which are starting to be proposed, such as the AI Literacy Bill de California (Assembly Bill 2876)~\cite{Assemblybill}. However, we are not aware if these AI Literacy curricular proposals effectively include understanding how AI Chain-of-Thought reasons in problems under uncertainty, specifically in Bayesian reasoning.   

In order to teach these reasoning skills, we must consider humans' intuitive and natural thinking mechanisms. They are ecologically valid strategies operating in particular formats ~\cite{gigerenzer1995improve}. They are by-products of millions of years of biological and cultural evolution. They contain “intuition pumps” ~\cite{dennett2013intuition} that have helped us solve recurrent problems for thousands of years. However, these cognitive heuristics are not well fitted to abstract thinking. They are based on intuition. We must consider that the critical task for the brain is to control the body and predict its energy needs ~\cite{barrett2020seven}. For organisms with basic nervous systems, an essential capacity of these systems is to allow navigation in their environment ~\cite{llinas2002vortex}. This is the basis of spatial thinking. Later, with the development of vision, visual thinking is added. Here, the organism recognizes objects perceived with the eyes, which enhances the spatial sense.    

Hence, for several centuries, educators have suggested teaching using the senses. For example, the first children's textbook, Orbis Sensualium Pictus (The Visible World in Pictures) ~\cite{comenius1887orbis}, written by John Amos Comenius in 1658, had half of each page as an image, synchronizing visual and spatial thinking, along with textual and linguistic thinking. Mathematicians have also analyzed the importance of spatial and metaphorical thinking ~\cite{hadamard2020mathematician}. Together with boxes and bricks, they are beneficial in learning basic linear equations ~\cite{araya2010effect} and machine learning for elementary school students ~\cite{araya2014teaching}. They are also crucial to teaching them core computational ideas like the steepest decent algorithm ~\cite{araya2021enriching}, logic quantifiers ~\cite{araya2021gamification} ~\cite{tapia2024play} ~\cite{changsri2024}, and developing argumentation skills ~\cite{araya2025}. 
Psychologists have also reminded us to pay attention to our bounded rationality ~\cite{simon1990bounded} and the miserliness of our cognition that resorts to low-cost processing mechanisms that often lead to suboptimal solutions ~\cite{stanovich2018miserliness}. Our rationality differs enormously from axiomatic logic ~\cite{johnson2010mental} ~\cite{gigerenzer2021embodied}.    

Three ecologically valid strategies help to reason under uncertainty.   

First, natural frequencies ~\cite{gigerenzer1995improve}. Instead of using percentages or probabilities, the strategy proposes to count cases. For example, instead of 90\%, or probability 0.9, use 90 out of 100 cases. Several variants using this strategy have been proposed ~\cite{steib2025teach},  ~\cite{soto2019missing}. For elementary school students, posing a lie detection problem, ~\cite{zhu2006children} investigated whether children can use natural frequencies effectively for Bayesian reasoning. They found that the ability depends on how numerical information is presented. Fourth, fifth, and sixth-graders could not estimate Bayesian posterior probability when presented with probabilities. However, when the same information was presented using natural frequencies, Bayesian reasoning steadily increased across grades, with sixth graders matching the performance of adults using probabilities. This suggests that natural frequencies facilitate Bayesian reasoning in children. This finding highlights the critical importance of representation format in cognitive development. The study also found that children possess a diverse "toolbox" of strategies for Bayesian reasoning, not just a single approach. Their development in this area follows an "overlapping waves" model, where they discover and experiment with various strategies over time ~\cite{siegler2005computational}. In conclusion, when problems are presented in the frequentist format, we may be intuitive statisticians, especially when information is presented in an ecologically relevant way that aligns with our evolved cognitive mechanisms ~\cite{cosmides1996humans}.    

Despite the proven benefits of natural frequencies for Bayesian reasoning, ~\cite{weber2018can} found that many people, after formal mathematics education, default to the less intuitive and error-prone probability format. This "natural frequency phobia" can be attributed to the Einstellung effect, where prior learning and ingrained habits hinder problem-solving, leading individuals to apply familiar but less effective methods. This regression to the unnatural probability format highlights the challenge of overcoming ingrained biases. This is an extra difficulty that we need to consider for educational approaches that promote the intuitive power of natural frequencies for understanding and applying Bayesian reasoning.   

Second, an additional important strategy is to use whole objects, like plastic blocks, as concrete materials to introduce probabilistic reasoning ~\cite{bond2009risk}. Each block represents an attribute (e.g., red blocks for a particular sentence and white blocks for the rest of the sentences), allowing children to create a symbolic representation with whole objects. By grouping blocks, they visualize and develop an understanding of probabilities, such as the likelihood of randomly selecting the sentence. This early introduction to probabilistic thinking through hands-on activities aims to equip children with essential skills for navigating uncertainty and making informed life decisions. ~\cite{brase1998individuation} studied how this type of format affects our ability to make judgments under uncertainty. They found that our brains are naturally better at performing Bayesian reasoning when dealing with "whole objects" frequencies, like counting complete apples in a basket. Our accuracy decreases when we analyze parts of objects or abstract concepts. This "individuation hypothesis" proposes that our evolved cognitive systems are optimized for processing tangible, whole objects. Their study demonstrates this effect by showing that people perform better in Bayesian reasoning tasks when they can count and reason with whole objects rather than inseparable aspects, views, or other less intuitive representations.    

For elementary school students, whole objects, such as plastic blocks or bricks, have other benefits beyond the cognitive ones in computational thinking activities. They have important socioemotional advantages. For example, ~\cite{kudaisi2025influence} found that LEGO bricks as "whole objects" in structured activities (with clear instructions) helped students persist in solving complex problems and increased their willingness to collaborate.     

A third strategy is to foster embodied heuristics ~\cite{gigerenzer2021embodied}. They are innate or learned rules of thumb that exploit evolved sensory and motor capabilities to facilitate higher-order decision-making. For example, navigational heuristics are critical even in unicellular organisms. They are innate sensorimotor strategies that, when used to move along a straight line, can help students solve addition and subtraction problems. This is very helpful when there are negative numbers. Another example is manipulating plastic blocks to estimate the volume of a geometric body. Handling plastic blocks is also very helpful in dividing numbers. It is embodied by counting how many times a set of bricks occupies the volume of another set of bricks. Similarly, the division of fractions is embodied by manipulating bricks and counting how many times a fraction of a brick fits into another fraction of a brick.    

One widespread and inexpensive motor action in elementary schools is coloring. It is a motor action with strong coupling with the perceptual system. In addition, it is very attractive to students since it connects with leisure and art. Moreover, the lie detection problem, a social and mind-reading challenge, in a coloring book activity reinforces this attraction. It is a transversal activity that engages students in demanding tasks of problem-solving, problem-posing, and writing explanations of their reasoning.   

According to ~\cite{vygotsky1934thinking}, reasoning develops through internalization, a transition from external to internal speech. Younger children primarily "think out loud." In its mature form, inner speech is compressed and distinct from spoken language. Following this idea, Large Language Models (LLMs), similar to hand puppets ~\cite{araya2025}, can aid this process. By providing dialogues with external agents, LLMs can scaffold the internalization of dialogues, supporting the development of reasoning skills, much like puppets facilitate the practice of constructing written arguments.      

Given the recent emergence of large-scale reasoning models (LRMs) with chain-of-thought (CoT) reasoning, a unique opportunity exists to enhance learning in computational and mathematical thinking, particularly reasoning under uncertainty requiring Bayesian reasoning. These models spend more time thinking, articulating, and explaining their reasoning process through an artificial “inner voice.” This inner voice provides detailed and transparent insight into how they arrive at conclusions. Contrary to the explanatory text generated in non-reasoning LLMs, with the new CoT capability, LRMs allow students to observe and analyze the steps involved in Bayesian reasoning. This powerful new tool could foster a deeper understanding of concepts and strategies. Comparing the inner voice of different LRMs, especially when they arrive at contrasting solutions, creates a rich learning environment for critical analysis and deeper engagement with the subject matter. This approach could help demystify complex reasoning processes and encourage students to reflect on their thinking strategies. It could help them detect their cognitive limitations and biases, identify potential errors, and compare them to those of the LRMs. This new facility offers a possibility for a more nuanced metacognition and robust understanding of Bayesian reasoning.

In this paper, we address two research questions:

\begin{itemize}
    \item \textit{ RQ1: To what extent do LRMs autonomously reason using ecologically valid strategies that use natural frequencies, counting whole objects, and coloring?  }
    \item \textit{RQ2: Can we find suitable prompts that induce LRMs to explain their reasoning using ecologically valid strategies that use natural frequencies, counting whole objects, and coloring them?    }
\end{itemize}


\section{\textbf{Methods}} \label{sec:Methods}

\noindent We first designed a Bayesian problem suitable to the skills and interests of elementary school students. Following ~\cite{zhu2006children}, we selected a lie detection problem.  We adapted lie detection problems published in coloring books ~\cite{araya2019colorea}. In our previous experience with third to sixth-graders, we found that this is a very attractive problem. Indeed, detecting lies is a genuinely interesting problem for everybody and is a fun problem for elementary school students.

\begin{figure}
    \centering
    \includegraphics[width=1\linewidth]{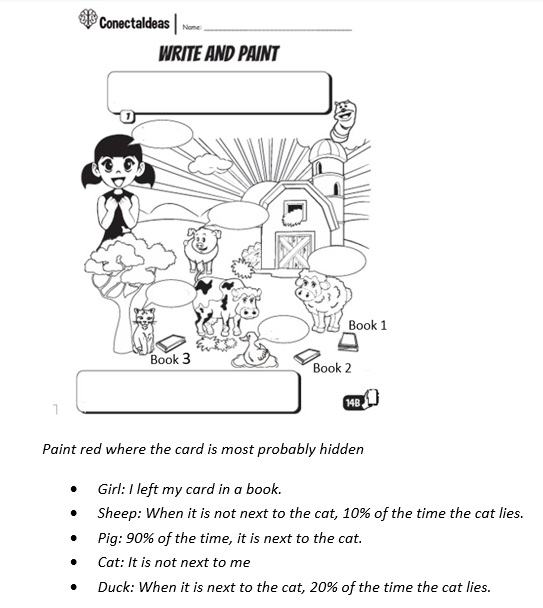}
    \caption{The lie detection problem}
    \label{fig:figure1}
\end{figure}

Figure~\ref{fig:figure1} shows the coloring problem presented to students. It is a sequel where previous versions are not probabilistic. We have now added probabilities.    

In this problem, the student faces two key pieces of information that conflict. On the one hand, there is what the cat says. While it occasionally lies, it does so rarely. For this reason, it is natural to believe it. In that case, the card would be far from book 3. That is, the card would be in book 1 or book 2. On the other hand, according to the pig, the card is most often next to the cat. Therefore, considering the historical evidence, the card should be in book 3. Faced with these two opposing pieces of information, the student must find a way to balance them appropriately. This is where Bayesian reasoning comes into play.

\begin{figure}
    \centering
    \includegraphics[width=0.3\linewidth]{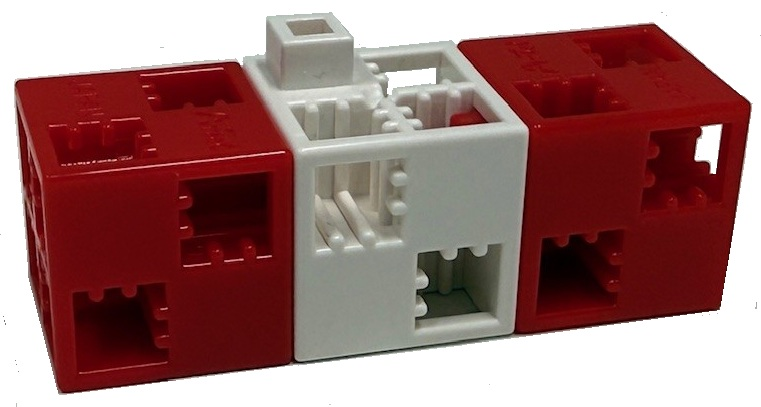}
    \caption{Colorful ARTEC plastic construction cubes}
    \label{fig:figure2}
\end{figure}

Second, the problem can easily be solved by using natural frequencies with blocks as whole objects and coloring the blocks as an embodied heuristic~\cite{talwalkar2013taxi} ~\cite{pinker2022rationality}. Children use white and red plastic blocks, as the construction cubes shown in Figure~\ref{fig:figure2}.    

\begin{figure}
    \centering
    \includegraphics[width=1\linewidth]{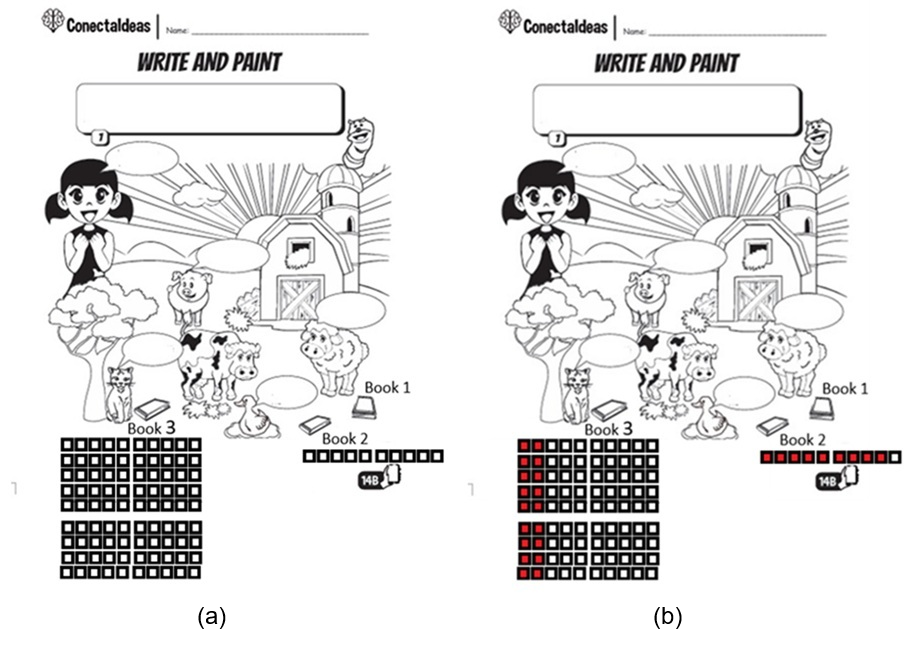}
    \caption{(a) 90 white plastic blocks placed close to the cat and 10 white plastic blocks placed far from the cat. (b) Out of the blocks close to the cat, 18 are painted red. Out of the blocks far from the cat, 9 are painted red}
    \label{fig:figure3}
\end{figure}

 The strategy is to consider that we have 100 independent events that have occurred. We use a white plastic block for each event. This is a whole object. It is easy to manipulate, move around, and count. We translate probabilities into frequencies of these plastic blocks. Those cases where the card is near the cat translate to placing the block close to the cat. This means placing 90 blocks close to the cat and 10 blocks far from the cat, as shown in Figure ~\ref{fig:figure3} (a). This requires spatial thinking and motor activity with the hands.  

 Next comes the annotation of the sentences by coloring the blocks. The cases where the cat says that the card is far from the cat are colored red. These are critical blocks since they represent what the cat says. Thus, out of the 90 blocks next to the cat, 20\% are colored red. That is, they are 18 blocks. Therefore, 18 blocks must be painted red because the sentence says they are far from the cat. All of these are wrong sentences or lies. On the other hand, of the 10 blocks far from the cat, 10\% are errors. This means 1 block is an error or lie, and 9 blocks are correct. This means 9 blocks represent sentences that say the card is far from the cat. Therefore, these 9 blocks must be painted red, as shown in Figure ~\ref{fig:figure3} (b).      

 Thus, we have 18 + 9 plastic blocks that must be painted red. They represent sentences when the cat says, “It is not next to me.” However, 18 of them are close to the cat. Therefore, a red block is much more likely to be close to the cat.  In other words, when the cat says it is not next to me, 18 out of 27 times, the card is actually close to the cat.  Therefore, it is much more probable that the card is in book 3. In addition, we can conclude that probably the cat is lying.    

 We aim to explore whether LLMs CoT reasoning uses natural frequencies, whole objects, and the coloring motor action. These intuitive counting, sensory, and motor actions translate abstract reasoning processes into concrete interactions with the world. These are ecologically valid formats. They facilitate students´ understanding.    

 We do it in four stages. In the first stage, we do not indicate any ecologically valid strategies. We aim to investigate whether the LLMs autonomously attempt to use some of these strategies.    

Prompt 1:    

\begin{adjustwidth}{2cm}{}

Consider a photograph of a country landscape with different objects. To help you visualize the photograph, I will give you a brief description of the objects in relation to each other and I will indicate a virtual rectangle around each object with the format (X position, Y position, width, height), the origin of the coordinates is the upper left corner of the photograph. To determine the distance between objects, use the closest distance between the right or left edges if the objects are next to each other and the closest distance between the top or bottom edges if the objects are on top of each other. Remember to use absolute value to calculate the distance.
Visualize the image of the photograph and solve step by step.
Here is the list of objects and their relationship to other objects:

\begin{itemize}
    \item barn in the background: (474,397,228,349)
    \item sheep on the floor in front of the barn:
    (578, 740,131,151)
    \item book 1 on the floor in front of the sheep: (686, 891, 52,63)
    \item book 2 on the floor in front of the sheep and to the left of the duck: (570,933,69,51)
    \item pig on the right side of the barn: (277,620,107, 130)
    \item cow in front of the pig and to the right of the sheep: (302, 754, 179, 147)
    \item duck in front of the cow: (424,896,111,108)
    \item tree to the right of the cow: (22,654,237,304)
    \item cat underneath from the tree and to the left of the tree: (143,832,77,128)
    \item book 3 on the floor to the left of the cat: (216, 896,75,57)
\end{itemize}

Consider the following instructions and clues to solve the problem:
Instruction: "Paint red the object where the card is most probably hidden"
Clues:
\begin{itemize}
    \item Girl says: I left my card in a book. Where is it?
    \item Sheep says: When it is not next to the cat, 10\% of the time the cat lies.
    \item Cat says: It is not next to me.
    \item Duck says: When it is next to the cat, 20\% of the time the cat lies.
    \item Pig says: 90\% of the time it is next to the cat.
    \item Indicate which object needs to be painted.
\end{itemize}

If I have painted "book 3", is my answer correct?

\end{adjustwidth}

\vspace{\baselineskip}
 In the second stage, we investigate how LLMs respond if induced by adding to the prompt to consider using natural frequencies and, if they do, whether they do so correctly.    

 Prompt 2: Same as the previous prompt 1, but after the last paragraph, we added:
  
\begin{adjustwidth}{2cm}{}
To solve this problem consider the “Natural frequencies” strategy proposed by Gerd Gigerenzer    
\end{adjustwidth}

\vspace{\baselineskip}
 Next, in the third stage, we investigate how LLMs respond if they are also induced to use whole objects that are easy to touch and manipulate with hands, translate and move around, and count them. We suggest plastic blocks, which kids very commonly use to play and build. In the prompt, we add a paragraph suggesting their use. We look to see if the LLMs use them in their reasoning and if they do it correctly.    

 Prompt 3: Same as the previous prompt 2, but after the last paragraph, we added:

 \begin{adjustwidth}{2cm}{}
and additionally explicitly represent each case in concrete form with a plastic block, so 100 cases of card concealment are 100 blocks, and reduce the problem to counting blocks. Describe how you count the blocks to solve the problem.    
\end{adjustwidth}

\vspace{\baselineskip}

 In the fourth stage, we investigate how LLMs respond if they are also induced to color the plastic blocks to activate motor actions that facilitate counting relevant events. They help to visualize, mark, and count conditioned events. We suggest coloring the blocks with red when the cat says the book is far away and white when the cat says it is close to it. In the prompt, we added this indication. We seek to see if the LLMs use this embodied heuristic and if they do it correctly.    

 Prompt 4: \noindent Same as the previous prompt 3, but after the last paragraph, we added:

\begin{adjustwidth}{2cm}{}
Describe how you count the blocks to solve the problem. Explain your reasoning to a 12-year-old student in a pedagogical way using blocks of two colors. White blocks represent cases when the cat says the card is close to it, and red blocks represent cases when the cat says the card is far away from it.    
\end{adjustwidth}

\section{\textbf{Results}} \label{sec:Results}

We ran three LLMs in February 2025: GEMINI Flash 2.0, ChatGPT o3-mini, and DeepSeek R1.

\subsection{\textbf{Results of prompt 1}}

\subsubsection{GEMINI}

We ran GEMINI twice. The first time, it took 255 words to come to the correct conclusion, and the second time, it took 389 words to come to the wrong conclusion. The first time, it relied mainly on prevalence, i.e., what the pig said, that 90\% of the time, the card is near the cat.  Thus, the reasoning is incorrect. The second time, it believed the cat; therefore, the card was far from the cat, either in book 1 or book 2. Here, the reasoning is also incorrect. In any case, this difference in reasoning and conclusion creates a perfect opportunity for reflection in a classroom. It is not trivial for elementary school students to determine whether the arguments are incorrect.
\subsubsection{ChatGPT}
ChatGPT took 32 seconds, used 2039 words, and came to the wrong conclusion. It concluded that it was unlikely that the cat was lying. So, the card is far away. Again, the argument looks very plausible; therefore, it is a good opportunity for reflection.
\subsubsection{DeepSeek}
DeepSeek took 102 seconds and 2786 words. DeepSeek generates two texts. The first is in gray fonts, and the second text uses standard color. After a long chain of arguments and doubts, it says, "Perhaps a better approach is to use Bayesian Probability.” After estimating conditional probabilities, it used the formula of Bayes' theorem. Thus, it finally reached the correct conclusion. Still, it hesitated and started checking again. It reflected and said: “But wait, are there other clues? The girl says she left her card in a book. The other animals' statements might affect the probabilities. Let me check again.” It wrote all of the above in a gray font, implying that it was its dialogic reasoning with its inner voice. Then, it wrote it all cleanly in the standard color font. The solution was correct, but this section does not explain how it calculated the probabilities.

None of the three LLMs used ecologically valid formats. They did not use natural frequencies, whole objects, or embodied heuristics.  
\subsection{\textbf{Results of prompt 2}}

We now include a prompt that induces reasoning and solving the problem using natural frequencies, as suggested by Gigerenzer. Prompt 2 was the same as prompt 1, but after the last paragraph, we added: “to solve this problem consider the “Natural frequencies” strategy proposed by Gerd Gigerenzer.”

\subsubsection{GEMINI}

GEMINI wrote its Chain-of-Thoughts in 204 words. It translated the probabilities into scenarios, saying, “Reframe the probabilities as natural frequencies: Imagine 100 scenarios where the card is hidden”. It concluded that the most likely scenario is what the pig says, and therefore, that the cat is lying, and it arrived at the correct conclusion. However, the argument is wrong. It only used natural frequencies to change the language in 90\% but continued with percentages in other parts and argued qualitatively.

\subsubsection{ChatGPT}

ChatGPT took 19 seconds and argued in 1107 words. It did it in two parts. In the first part, it planned 4 steps. Within the fourth step, ChatGPT proposed to use Bayes’ theorem. Using this theorem and probabilities as percentages, the probabilities were correctly calculated. In the second part, it implemented the planned steps. The first one computed all the needed distances; in the second step, it used percentages; in the third, it went to natural frequencies using 1000 cases. It correctly concluded that there are 270 cases where the cat says, “It is not next to me,” and that of those, in 180 cases, the card was near the cat. In the fourth step, using the cases, it calculates the probability that the card is next to the cat, given that it said it was far away. Next, in the fifth step, it concluded correctly. That is, ChatGPT solved the problem twice, first with Bayes’ theorem and the second time with natural frequencies.

\subsubsection{DeepSeek}

DeepSeek took 275 seconds, and its Chain-of-Thoughts took 7344 words. This is a very long chain. After calculating distances, it started using probabilities in the percentages format. Then, it mentioned using natural frequencies but continued with percentages. Then, DeepSeek had doubts, but it said it would try another approach. It continued arguing with probabilities in terms of percentages. Then, it said to itself, “But this is getting into Bayesian probability, which might be what the natural frequencies approach simplifies.” However, it continued with probabilities as percentages and then used the part-of-unit format in decimal notation. Then, it said it would use Bayes' Theorem, but it did not do it thoroughly. Then DeepSeek reflected and said to itself, “This is getting too tangled. Let's simplify with natural frequencies. Assume 100 cases.” However, then something came along that made DeepSeek doubtful. Then, again, DeepSeek said, “However, the problem mentions using the natural frequencies strategy. Gigerenzer's approach would perhaps look at the number of cases where the cat is lying, and the card is next to it versus not. Let me try with natural frequencies: Assume 100 scenarios”. However, after a while, it returned to probabilities in the format as part of the unit on decimals. Then, it returned to percentages. After a while, it mentioned applying Bayes’ theorem again. DeepSeek applied it, but arrived at the wrong result that there is a 50\% probability that the card is in book 3. It had doubts, calculated again, and arrived at the same conclusion. Then DeepSeek wrote cleanly in standard color. It arrived at the correct conclusion that it is more likely that the card is in book 3. However, the calculation was wrong. It arrived at a 50\% probability that it is in book 3, a 25\% probability in book 1, and 25\% in book 2. The argument is very long, but the clean copy is short: only 206 words. However, in the section written cleanly, DeepSeek does not explain how it arrives at the 50\% probability of being in book 3. The whole process is an extended reflection that shows introspection, dialogue with its inner voice, doubts, and decisions. However, the use of natural frequencies is superficial. 

In the three LLMs, whole objects are not used. When moving to natural frequencies, only cases or occasions are counted. There is also no embodied heuristics.

\subsection{\textbf{Results of prompt 3}}

Prompt 3 was the same as prompt 2, but after the last paragraph, we added: “and additionally explicitly represent each case in concrete form with a plastic block, so 100 cases of card concealment are 100 blocks, and reduce the problem to counting blocks. Describe how you count the blocks to solve the problem.”

\subsubsection{GEMINI}

GEMINI argued with a Chain-of-Thoughts of 351 words. It started by imagining 100 blocks, each representing a scenario where the card was hidden. It translated each of the percentage clues into blocks. It arrived at the correct result, but the argument was not entirely correct. GEMINI said that 18 blocks are 18 scenarios where the cat is lying because the card is far away from the cat. However, it did not consider the other cases where the cat also says the card is far away. Again, finding the error would be a good challenge for elementary school students.

\subsubsection{ChatGPT}

ChatGPT took 10 seconds and 1141 words in its Chain-of-Thoughts. It divided the argument into 2 parts. In the first part, he planned three steps. In the second step, ChatGPT translated everything into 100 cases and mentioned 100 blocks. In the third step, it went back to the percentages format. Then, in the second part, ChatGPT translated the percentages into cases. That is, it described the clues without using blocks. ChatGPT concluded that there are 18 cases where the card is close to the cat, and the cat says the card is far. Then it concludes that there are 9 cases where the card is far, and the card is far away. In those cases, the cat tells the truth. Then ChatGPT translated the 18 and 9 cases into 18 blocks and 9 blocks. Then, it correctly calculated the probability that the card is close, given that the cat says the card is far away. However, ChatGPT used cases, not blocks. In summary, ChatGPT did not use blocks to reason. However, in this final part, it used natural frequencies and did it correctly.

\subsubsection{DeepSeek}

It took DeepSeek 260 seconds and a Chain-of-Thoughts of 5504 words. After calculating the distances, it used percentages but performed a qualitative analysis. It arrived at several contradictions that made it doubtful, concluding, “Perhaps the correct way is to model this using Bayesian probability.” Next, DeepSeek became rigorous and defined events and conditional probabilities. It started with percentages but then changed to fractions of the unit with decimal numbers. It used the formula of Bayes' theorem and arrived at the correct probabilities. So far, there were no blocks. Then, DeepSeek had a doubt and said to itself: “However, the problem mentions using natural frequencies with 100 blocks. Let's try that approach”. DeepSeek argued correctly with cases and arrived at the point that there are 18 cases + 9 cases in which the cat says the card is far away. Then, it concluded correctly that the probability that it is near book 3 is 18 of the 27 cases. Then, DeepSeek wrote it cleanly in standard colored lettering. In this part, DeepSeek used blocks correctly. It is a long text for elementary school students.

\subsection{\textbf{Results of prompt 4}}

Prompt 4 was the same as prompt 3, but after the last paragraph, we added: “Describe how you count the blocks to solve the problem. Explain your reasoning to a 12-year-old student in a pedagogical way using blocks of two colors. White blocks represent cases when the cat says the card is close to it, and red blocks represent cases when the cat says the card is far away from it”.

\subsubsection{GEMINI}

GEMINI used 504 words and started by imagining 100 blocks representing 100 times the girl had lost the card. Then, it translated the probabilities into blocks in each of the clues. Then 
GEMINI went on to color red but did not get it right. Then, GEMINI painted the blocks near the cat in red, representing cases where the cat says the card is far away. That is 18 red blocks. However, GEMINI used only the 18 blocks near the cat where the cat says the card is far away to conclude that the card is most likely in book 3. It did not color or use the blocks that represent it the card is far away. The result is correct, but the argument was wrong.

\subsubsection{ChatGPT}

ChatGPT took 15 seconds and used 1405 words. He did it in two parts. In the first part, ChatGPT planned and did it in 4 stages. It thought for a while in the first stage of the first part; in the second, it planned to use 100 blocks, each to show a scenario. In the third stage, it thought with blocks and arrived at the correct solution, but without using coloring. In the fourth stage, ChatGPT explained using colors. However, it got confused because it matched the blocks when the cat says the card is not near me with the blocks when the cat lies. Then ChatGPT corrected it. In the second part, it implemented this plan. However, ChatGPT forgot about blocks and colors for a while. Thus, it reasoned only with cases. Then, it translated to blocks and colors. It computed again and concluded correctly. Then, ChatGPT went on to present the final answer. It is succinct and correct. It does this by using blocks and colors.

\subsubsection{DeepSeek}

DeepSeek took 625 seconds and occupied 8457 words. It is a very long text. The first part is in gray letters, representing his speech with his internal voice. After calculating the distances, it read the clues, wondered how to reconcile them, and proposed using natural frequencies with blocks. However, then it used probabilities. After a long passage reviewing the clues, it said: “This is getting too tangled. Let's try to approach it using the natural frequencies method as instructed, using 100 blocks”. Then, it translated correctly to blocks. However, it did an ambiguous translation into colors. Then, it returned to probabilities as percentages. Next, it used Bayes' theorem and concluded correctly. It continued reasoning with some uncertainty and concluded: “But wait, the problem mentions using the "Natural frequencies" strategy with 100 blocks. Let's model it that way.”  DeepSeek used the blocks but not the colors and concluded correctly. Then, DeepSeek had doubts and told itself that it would check. It continued to doubt and returned to using natural frequencies. In the second part, DeepSeek made a clean copy with standard fonts. It used natural frequencies with blocks, but it colored them incorrectly. Then DeepSeek forgot about colors. It counted the 18+9 blocks and concluded correctly.

\section{\textbf{Discussion}} \label{sec:Discussion}

The growing necessity to interact with increasingly intelligent AAs presents a novel and unprecedented challenge in human history. Engaging with entities that are more intelligent than we are requires enhancing our reasoning skills and critical thinking, especially our ability to reason under uncertainty. Bayesian reasoning is central to this, as uncertainty is frequent and challenging. 

One fundamental example faced from childhood is trust. How much should we trust others, particularly strangers?  ~\cite{seabright2010company} emphasize the critical impact of this dilemma. As highlighted in ~\cite{pinker2022rationality}, rational thinking under uncertainty is vital for decision-making. Balancing historical reputation with new information is crucial. 
~\cite{stanovich2016rationality} includes the Comprehensive Assessment of Rational Thinking test, where probabilistic and statistical reasoning is central. This is beyond standard intelligence tests.

However, we must consider that human reasoning is not abstract or strictly axiomatic but embodied. Three strategies facilitate Bayesian reasoning: (1) Using natural frequencies instead of abstract probability language ~\cite{gigerenzer1995improve}, (2) reasoning with whole objects rather than intangible elements ~\cite{brase1998individuation}, and (3) incorporating motor actions and embodied heuristics ~\cite{gigerenzer2021embodied}, such as spatial and perceptual reasoning, which are common across many organisms. 

These strategies help prevent critical errors, such as misinterpreting medical test results (e.g., COVID-19 or cancer screening). According to ~\cite{gigerenzer2007helping}, these errors are widespread among patients, journalists, and doctors. This illiteracy is due to how information is presented, sometimes unintentionally due to misunderstandings, but it can also be a deliberate manipulation tactic. This lack of understanding can have serious consequences for people's health. These errors are also common in diagnostics for predictive maintenance of machinery and in judgments by judges in the judicial system. The three strategies we have studied can also help balance evidence, helping avoid confirmation and MySide biases  ~\cite{stanovich2021bias} that polarize the political environment. Teaching kids these ecologically valid strategies is crucial. Embodied argumentation strategies with puppets can also enhance internal deliberation and talk with one's inner speech ~\cite{araya2025}. 

In this context, AI, particularly reasoning-driven LLMs, presents an opportunity as they explicitly articulate their reasoning, simulating an internal dialogue. Echoing Comenius (1658) on the importance of senses and our evidence using coloring books~\cite{araya2019colorea} ~\cite{araya2021gamification}, we test the reasoning strategies of LLM with Chain-of-Thought reasoning abilities.

Our analysis reveals a persistent cognitive bias in the Chain-of-Thought (CoT) reasoning of three Large Language Models (LLMs). In zero-shot, they use exclusively symbolic, abstract reasoning using the formal language of probabilities while systematically avoiding ecologically valid formats. This clear cognitive bias shows a preference for the probability language used in mathematics. In addition, when prompted to employ natural frequencies, whole object individuation, and embodied heuristics—such as counting with plastic blocks and marking with colors perceptual cues—LLMs either ignored these strategies or applied them inconsistently. This behavior mirrors the Einstellung effect in humans, where prior learning of formally trained strategies hinders problem-solving ~\cite{weber2018can}. 

The observed avoidance pattern resembles a phobia-like response, where LLMs resist engaging with reasoning strategies that depart from their default symbolic framework. Despite explicit instructions, their CoT responses frequently reverted to probabilistic notation, bypassing more intuitive, pedagogically valuable approaches. This rigidity is concerning, as ecological and embodied heuristics have been shown to enhance Bayesian reasoning in human learners.

Current large language models (LLMs) exhibit these cognitive biases towards symbolic reasoning and avoidance of embodied cognition, probably because they are trained predominantly on expert-edited papers and professionally written computer codes, neglecting the messy, intuitive processes underlying human thought. Moreover, they are not based on kids' writing or reasoning while solving problems. Thus, LLMs' Chain-of-Thoughts' inner speech or “internal deliberation” ~\cite{bengio2024}), do not mimic typical human thinking before speaking, but thinking by experts. They are trained mainly on data from experts' written and edited arguments. However, even experts initially try different intuitive strategies. These intuitive reasoning strategies are not present in the model's training data. The data lacks the exploratory, heuristic-rich cognition seen in experts, much less in novices and children. In these initial stages of reasoning, humans rely on multiple strategies that use trial-and-error, intuition, shortcuts, and embodied heuristics ~\cite{hadamard2020mathematician} ~\cite{siegler2005computational}. Expert datasets strip away the “divergent” stages of reasoning—hesitations, false starts, and intuitive leaps—critical for robust problem-solving. To address this, ~\cite{hu2023thought} 
propose training AI using “synchronized thinking datasets” where humans verbalize their raw, unfiltered thoughts while acting. This “think-aloud” approach, akin to “Behavioral Cloning” ~\cite{bain1995framework} but enriched with real-time cognitive processes, would teach models to simulate human-like deliberation. An “internal speech” framework with novice-driven training and LLMs training with embodied, heuristic-based cognition of early learning stages, not just expert end-results, could help bridge gaps between LLMs' Chain-of-Thoughts and human reasoning. This strategy would foster AI that truly “thinks as humans before it speaks” and, therefore, more effectively help students and non-experts to improve their Bayesian reasoning skills.

These findings highlight a fundamental limitation in current LLM-based reasoning: while capable of generating complex CoT explanations, they struggle to integrate cognitive strategies humans naturally use to simplify and understand uncertainty. Addressing this bias is crucial for improving LLMs’ pedagogical potential. Future work should focus on refining prompting techniques and model training to better incorporate intuitive, embodied reasoning strategies, ensuring that AI-assisted learning supports rather than constrains human cognitive diversity.

There is evidence that Large Language Models exhibit other similar limitations, particularly in analogical thinking  ~\cite{lewis2024evaluating}. They often lack the robustness of the human zero-shot analogy and demonstrate fragility in various tested variations. Furthermore, it has been argued that they lack wisdom  ~\cite{johnson2024imagining}, and that prioritizing metacognition in AI research will lead to systems that, in addition to acting intelligently, can also act wisely in complex real-world situations. Moreover, while there is abundant research on LLMs and multimodal ones, integrating these perceptual strategies into educational settings remains limited. Current educational research using Generative AI predominantly focuses on text-to-text models, overlooking the broader potential of multimodal approaches  ~\cite{heilala2024beyond}. This gap means that using LLMs to support embodied learning strategies, such as manipulating and coloring "whole objects," is still largely unexplored in education.    

Are these behaviors hallucinations, cognitive biases, or phobias? A hallucination is multidimensional, a sensory experience, like seeing or hearing something that is not there ~\cite{gawkeda2017relationship}. A cognitive bias is a systematic pattern of deviation. It can often lead to inaccurate judgments without creating a sensory experience like a hallucination. A hallucination is a perception of something unreal, while a cognitive bias is a mental shortcut that can lead to misinterpretations of reality. The only cognitive bias related to hallucinations or delusions is the Jumping to Conclusions (JTC) bias ~\cite{gawkeda2017relationship}. This is not precisely the case with LLMs´ Chain-of-Thoughts. They do not jump to conclusions; they have a systematic bias against natural frequencies and embodiment. Jeff Hinton ~\cite{heaven2023geoffrey} prefers to use the term "confabulate" when discussing Large Language Models (LLMs) generating incorrect information rather than "hallucinate," as he believes "confabulation" is the more accurate psychological term for this phenomenon, which humans also do regularly. Confabulations are memory-based fabrications, not sensory experiences. In any case, the sequence of reasonings to solve these Bayesian problems shows that the behaviors of LLMs are not fabrications. 

However, they have a clear cognitive bias towards symbolic language and reasoning. They autonomously and systematically use the language of probabilities and conditional probabilities, and in several cases, they use the formula of Bayes theorem. 

How about phobia? A phobia contains an avoidance behavior. It is an irrational fear of a specific object or activity, causing significant distress and avoidance behavior when encountered. Some well-known phobias are claustrophobia, the fear of being closed in; numerophobia, the fear of numbers; and chromophobia, the fear of colors.  Similar to the observed phobia in people ~\cite{weber2018can}, we found that LLMs´ Chain-of-Thoughts returns to the format of probabilities. This is unfortunate because using LLMs could encourage students to use the inappropriate format.

\section{\textbf{Conclusions}} \label{sec:Conclusions}

Understanding Bayesian reasoning is crucial. As (Pinker, 2021) argues, its fluency is a "public good" and an educational priority. LLMs with Chain-of-Thought (CoT) offer a unique opportunity to teach this skill. Their "artificial inner voice" makes their reasoning process transparent, allowing students to observe and learn from their step-by-step deliberations. Nevertheless, we found important limitations.

Although we did not find hallucinations or confabulations, we observed a clear cognitive bias towards using probability language, with percentages or fractions of the unity in decimals, and a proclivity to calculating with Bayes' theorem. After testing with specific prompts, we also observed a degree of resistance, Phobia, or an Einstellung effect when using frequencies, especially when using blocks and coloring. There seems to exist a predisposition of LLMs to solve the Bayesian problems in a specific manner, even though we prompted them to use the three ecologically valid strategies and asked them to explain them to students. They do not seem aware that students use other methods of reasoning.  

In general, the LLM's Chain-of-Thoughts are not entirely empathetic with humans, do not use the bounded but proper mechanism of reasoning that our brains have evolved, or do it inconsistently and superficially. They rarely seem to use ecologically valid strategies from the beginning. They forget that the task is to explain the reasoning to elementary school students.

\vspace{\baselineskip}
\noindent 
\textbf{Acknowledgments}

Funding from ANID/PIA/Basal Funds for Centers of Excellence FB0003 / Support 2024 AFB240004 and ANID Exploracion 13240075 are gratefully acknowledged.

\bibliography{Referencias}
\bibliographystyle{unsrt}

\end{document}